# Segmentation en phrases : ouvrez les guillemets sans perdre le fil

Sandrine Ollinger[1], Denis Maurel[2]

[1]ATILF, Université de Lorraine, CNRS – Sandrine.Ollinger@atilf.fr

[2]Université de Tours, Lifat – Denis.Maurel@univ-tours.fr

## Abstract

This paper presents a graph cascade for sentence segmentation of XML documents. Our proposal offers sentences inside sentences for cases introduced by quotation marks and hyphens, and also pays particular attention to situations involving incises introduced by parentheses and lists introduced by colons. We present how the tool works and compare the results obtained with those available in 2019 on the same dataset, together with an evaluation of the system's performance on a test corpus.

## Résumé

Cet article présente une cascade de graphes pour la segmentation en phrases de documents XML. Notre proposition prévoit une inclusion de phrases pour les cas introduits par des guillemets et tirets et porte également une attention particulière aux situations d'incises introduites par des parenthèses et des listes introduites par des deux-points. Nous présentons le fonctionnement de l'outil et comparons les résultats obtenus à ceux disponibles en 2019 sur le même jeu de données, ainsi qu'une évaluation des performances du système sur un corpus test.

**Mots clés :** segmentation ; phrases ; inclusion ; Unitex ; cascades de graphes.

## 1. Introduction

La segmentation en phrases est une étape courante de prétraitement des données, qui permet leur exploitation dans des concordanciers avancés pour des recherches en contextes restreints[1]. Elle peut également être exploitée dans le cadre de comparaisons statistiques de contextes ou de traitements automatiques.

Nous nous intéressons à son exploitation dans le cadre de l'étude du lexique et de l'instrumentation de l'activité lexicographique. Dans cette perspective, nous avons souhaité fournir une version segmentée lors de la mise à disposition du corpus de citations lexicographiques BEL-RL-fr (Ollinger, 2023). Nous avons alors testé différents outils — nltk.sent_tokenize (Kiss et Strunk, 2006) ; Talismane 5.2.0 (Urieli, 2013) ; cascade CasEN développée sur la plateforme Unitex/GramLab (Maurel et al., 2011). Ils ont tous retourné des résultats insatisfaisants, provoquant des ruptures de relations syntaxiques entre éléments lexicaux, principalement dans le cas des phrases comportant des guillemets. Il nous est alors apparu que la notion de phrases secondaires incluses dans une phrase principale manquait pour obtenir un résultat convenable.

---

[1] On pense ici notamment à la syntaxe des requête CQL *within s*.

Les trois figures ci-dessous illustrent les résultats obtenus et attendus pour la citation (1), issue du corpus BEL-RL-fr. Tandis que la segmentation proposée pourrait amener à identifier un emploi absolu du verbe *dire[Quelqu'un dit quelque chose à quelqu'un]*, la segmentation attendue donne accès à son second argument.

(1.1)  Néné me tend un godet, l'air vaguement contrarié. « Kestu fous avec le flingue ? C'est l'ouverture d'la chasse ou quoi ? » « ça s'pourrait... » On trinque et puis y m'dit : « J'la nettoye. Ces machins-là si c'est pas entretenu ça risque de te péter à la gueule ! »

(1) Néné me tend un godet, l'air vaguement contrarié.
(2) « Kestu fous avec le flingue ?
(3) C'est l'ouverture de la chasse ou quoi ?
(4) » « ça s'pourrait…
(5) » On trinque et puis y m'dit :
(6) « J'la nettoye.
(7) Ces machins-là si c'est pas entretenu ça risque de te péter à la gueule !
(8) »

*Segmentation en 8 phrases proposée par les trois outils.*

(1) Néné me tend un godet, l'air vaguement contrarié.
(2) « Kestu fous avec le flingue ?
(3) C'est l'ouverture de la chasse ou quoi ? »
(4) « ça s'pourrait… »
(5) On trinque et puis y m'dit : « J'la nettoye.  Ces machins-là si c'est pas entretenu ça risque de te péter à la gueule ! »

*Segmentation en 5 phrases principales attendue.*

(5.1) « J'la nettoye.
(5.2) Ces machins-là si c'est pas entretenu ça risque de te péter à la gueule ! »

*Segmentation en 2 phrases secondaires du discours rapporté contenu dans la dernière phrase.*

Après avoir discuté des problèmes rencontrés, une collaboration s'est mise en place entre nous autour du corpus BEL-RL-fr et de la création d'une cascade de graphes Unitex, CasFin. Après une brève présentation des données sur lesquelles nous avons travaillé, nous présenterons dans la section 3 le fonctionnement du modèle de segmentation mis au point au cours de cette collaboration. Nous nous attarderons ensuite à comparer la segmentation proposée par ce nouveau modèle à celle fournie par CasEN fin 2019. Enfin, avant de conclure, la section 5 proposera une évaluation chiffrée des performances obtenues.

## 2. Données

Le BEL-RL-fr, dans sa version 1, est un fichier XML encodé en UTF-8. Il comporte 29 892 citations pour 1 033 574 mots[2]. Après avoir discuté autour de quelques cas particuliers rencontrés lors des tests de segmentation, nous avons scindé le corpus en deux parties : l'une consacrée au développement, comportant 19 892 citations et l'autre réservée à l'évaluation, comportant 10 000 citations. Pour enrichir nos données de développement, nous avons également eu recours aux citations lexicographiques du Trésor de la Langue Française Informatisé[3] contenant des guillemets. Ces dernières ne sont cependant pas prises en compte dans les résultats présentés ici.

La comparaison systématique des modèles de fin 2019 et de 2024 a été réalisée sur un sous-ensemble de 691 citations extraites de la partie développement de notre corpus. Il s'agit de citations comportant des guillemets, des parenthèses, des tirets ou des crochets. L'évaluation chiffrée proposée dans la section 5 a été réalisée sur un sous-ensemble de 3 000 citations extraites de la partie évaluation de notre corpus.

## 3. Fonctionnement

La cascade de graphes CasEN, créée pour la reconnaissance des entités nommées (Friburger et Maurel, 2004), commence par une prise en compte des balises XML et une segmentation en phrases, fortement inspirée du graphe proposé dans la distribution d'Unitex (Friburger et al., 2000). Nous en avons extrait les graphes concernés et nous les avons placés dans une nouvelle cascade, CasFin. La cascade ainsi réalisée marquait les débuts et fins de phrase, mais sans tenir compte des inclusions. Deux problèmes principaux se posaient : l'étude du contexte du caractère deux-points et les textes entre guillemets, parenthèses ou crochets.

### *3.1. Le contexte du caractère deux-points*

Dans CasEN, le caractère deux-points, suivi d'un mot commençant par une majuscule, est considéré comme une fin de phrase. Or il s'agit parfois simplement du complément de la phrase de départ, ce qui interdit la segmentation à cet endroit. Par exemple, dans notre corpus :

(3.1.1) Quand à Vincennes une poignée de femmes a levé l'étendard de la révolte, des gauchistes ont envahi la salle en criant : « Le pouvoir est au bout du phallus. »

\<s\>Quand à Vincennes une poignée de femmes a levé l'étendard de la révolte, des gauchistes ont envahi la salle en criant **:\</s\>**
**\<s\>"** Le pouvoir est au bout du phallus.\</s\> "

*Segmentation de la citation (3.1.1) obtenue par CasEN*

---

[2] Le nombre de mots a été calculé à l'aide de la plateforme TXM.

[3] Au moment de la rédaction de ce document, il n'existe pas encore de version XML diffusée du TLFi.

\<s\>Quand à Vincennes une poignée de femmes a levé l'étendard de la révolte, des gauchistes ont envahi la salle en criant **: "**
   **\<s\>**Le pouvoir est au bout du phallus.**\</s\>**
"\</s\>

*Segmentation de la citation (3.1.1) obtenue en mars 2024*

Pour cela, nous inférons que la présence d'un indicateur dans la partie de la phrase précédent le caractère *deux-points* rend fortement probable la poursuite de la même phrase après ce caractère. Par exemple, dans la phrase ci-dessus c'est le verbe *crier*. Cet indicateur peut être un verbe (*demander*, *insister*, *répondre*…), un nom, souvent déverbal (*ajout*, *chuchotement*, *explication*…), mais pas seulement (*article*, *bouche*, *secret*, *ultimatum*…) ou encore un présentatif (*voici*, *voilà*, *c'est*…), un adverbe (*Enfin*, *Ensuite*, *Finalement*…), une locution (*jouer un tour*, *voir clair*, *rendre hommage*…) ou encore la présence d'un nom propre. Nous avons donc ajouté à la cascade un graphe reconnaissant ces indicateurs dans un certain contexte.

Un graphe ajouté aux précédents découpe la phrase en segments de texte, en ponctuations et en indicateurs. Ainsi, notre exemple est partagé en trois segments et un indicateur : {Quand à Vincennes une poignée de femmes a levé l'étendard de la révolte,.Segment}, {des gauchistes ont envahi la salle en,.Segment} {criant,.Indicateur} :\</s\> " \<s\>{Le pouvoir est au bout du phallus,.Segment}. "\</s\>\</p\>Par contre, dans l'exemple suivant, la segmentation au caractère deux-points est correcte :

(3.1.2) Je suis loin d'avoir la vérité absolue, mais voici quelques pistes : Au niveau national, priorité doit être donnée à la voiture propre.

\<s\>Je suis loin d'avoir la vérité absolue, mais voici quelques pistes :**\</s\>**
\<s\>Au niveau national, priorité doit être donnée à la voiture propre.\</s\>

*Segmentation correcte de la citation (3.1.2) obtenue par CasEN*

### *3.2. Les guillemets, parenthèses ou crochets*

Prenons l'exemple suivant :

(3.2)   Vous accablez tous ce pauvre père qui n'a rien à voir dans l'histoire et vous vous arrêtez à la seule version que propose ce chiffon (parce que oui ! Cet article est un chiffon !) sans aller voir plus loin que le bout de votre nez parce que vous êtes dépourvu de tout sens critique !

\<s\>Vous accablez tous ce pauvre père qui n'a rien à voir dans l'histoire et vous vous arrêtez à la seule version que propose ce chiffon **(**parce que oui **!\</s\>**
   **\<s\>**Cet article est un chiffon !**)** sans aller voir plus loin que le bout de votre nez parce que vous êtes dépourvu de tout sens critique !\</s\>

*Segmentation de la citation (3.2) obtenue par CasEN*

L'absence d'une majuscule sur le mot *parce* crée une confusion : le segment *parce que oui !* est-il une phrase ? La présence d'une phrase dans la suite de la parenthèse nous amène à penser que oui… Son absence nous ferait plutôt pencher pour le contraire… Nous proposons donc les segmentations suivantes :

\<s\>Vous accablez tous ce pauvre père qui n'a rien à voir dans l'histoire et vous vous arrêtez à la seule version que propose ce chiffon **(**
   **\<s\>**parce que oui **!\</s\>**
   **\<s\>**Cet article est un chiffon **!\</s\>)**
sans aller voir plus loin que le bout de votre nez parce que vous êtes dépourvu de tout sens critique !\</s\>

*Segmentation de la citation (3.2) obtenue en mars 2024*

\<s\>Vous accablez tous ce pauvre père qui n'a rien à voir dans l'histoire et vous vous à la seule version que propose ce chiffon (parce que oui !) sans aller voir plus loin que le bout de votre nez parce que vous êtes dépourvu de tout sens critique !\</s\>

*Segmentation de la citation tronquée (3.2) obtenue en mars 2024*

## 4. Comparaison des modèles de 2019 et 2024

Les résultats insatisfaisants qui ont motivé le développement d'un nouveau modèle de segmentation en phrases, s'ils nous ont sauté aux yeux lors des tests réalisés en 2019, n'en sont pas moins un problème de faible étendue, comme on le voit dans le tableau ci-dessous.

| Sans changement | Amélioration totale | Alternative sans gain ni perte | Amélioration partielle | Problème persistant | Nouveau problème |
|---|---|---|---|---|---|
| 491 ; 71,5% | 90 ; 13% | 74 ; 10,5% | 16 ; 2,33% | 9 ;1,33% | 9 ;1,33% |

*Répartition des différences observées entre 2019 et 2024.*

En réalisant une comparaison systématique des différences de segmentation obtenue sur 691 citations avec la version CasFin de 2019 et celle à laquelle nous avons abouti, nous constatons que, dans 71,5 % des cas, le traitement est inchangé et correct. Dans 13 % des cas, les problèmes qui existaient dans la version 2019 ont été entièrement réglés à l'aide de la version 2024 : nous les classons dans la catégorie **amélioration totale**. Nous pouvons ajouter à cet ensemble de bons résultats, les 10,5 % de cas d'**alternative sans gain ni perte** et ainsi considérer que 95 % des segmentations obtenues sont entièrement satisfaisantes.

Dans les 5 % de cas restant, 16 correspondent à la catégorie des **améliorations partielles**, 9 à celle des **problèmes persistants** et 9 à celle des **nouveaux problèmes**. La suite de cette section illustre rapidement chacune de ces catégories, dont vous trouverez une présentation plus détaillée dans la documentation jointe à la distribution de notre modèle de segmentation.

### *4.1. Amélioration totale*

Les améliorations totales correspondent dans leur immense majorité aux phénomènes de phrases incluses qui nous avaient initialement interpelés. Elles concernent des cas de dialogues introduits par des guillemets, de dialogues introduits par des tirets, d'incises introduites par des parenthèses ou de listes introduites par des deux-points. Elles concernent également quelques points d'initiale anciennement pris pour des points de fin de phrase et une meilleure gestion des points non suivis d'espace. La citation (4.1) illustre une telle amélioration. La nouvelle segmentation obtenue délimite proprement le deuxième argument du verbe *déclarer[Quelqu'un déclare quelque chose à quelqu'un]*.

(4.1) « J'ai téléphoné à midi. Les onze cylindres spéciaux arrivaient par courrier exprès dès le lendemain matin », déclare-t-il, apparemment encore très satisfait du service de nuit rapide.

\<s\>" J'ai téléphoné à midi.\</s\>
\<s\>Les onze cylindres spéciaux arrivaient par courrier exprès dès le lendemain matin ", déclare-t-il, apparemment encore très satisfait du service de nuit rapide.\</s\>

*Segmentation de la citation (4.1) obtenue fin 2019.*

\<s\>"
　\<s\>J'ai téléphoné à midi.\</s\>
　\<s\>Les onze cylindres spéciaux arrivaient par courrier exprès dès le lendemain "\</s\>, déclare-t-il, apparemment encore très satisfait du service de nuit rapide.\</s\>

*Segmentation de la citation (4.1) obtenue en mars 2024.*

## *4.2. Alternative*

Le traitement des cas problématiques a eu des répercussions sur le traitement de cas que nous n'avions pas considérés comme étant mal segmentés en 2019. L'introduction de la notion de phrase englobante crée des situations qui peuvent sembler artificielles dans lesquelles des phrases se trouvent regroupées en bloc. Il peut s'agir de bloc de dialogue, dont l'étrangeté s'accroît lorsque le bloc est constitué d'une seule phrase. Des situations similaires sont observées avec les parenthèses. On peut imaginer que les utilisateurs gênés par ses regroupements pourront réaliser un post-traitement pour s'en défaire. Dans d'autres cas, l'alternative consiste uniquement en l'inclusion ou non d'un signe de ponctuation dans la phrase, comme pour les tirets.

(4.2) La fête avec ses manèges, ses stands de tir, de confiseries et de barbe à papa. « J'ai passé la soirée sur le Taïga Jet confiait Joffrey, 14 ans, c'est super. Il y a de l'ambiance, tous les ans j'y vais et à chaque fois c'est au top ! » Quant à Kévin, 8 ans, lui, il s'essayait près de sa maman à la machine à pinces.

\<s\>La fête avec ses manèges, ses stands de tir, de confiseries et de barbe à papa.\</s\>
\<s\>" J'ai passé la soirée sur le Taïga Jet confiait Joffrey, 14 ans, c'est super.\</s\>
\<s\>Il y a de l'ambiance, tous les ans j'y vais et à chaque fois c'est au top!\</s\>
\<s\>" Quant à Kévin, 8 ans, lui, il s'essayait près de sa maman à la machine à pinces.\</s\>

*Segmentation de la citation (4.2) obtenue fin 2019.*

\<s\>La fête avec ses manèges, ses stands de tir, de confiseries et de barbe à papa.\</s\>
\<s\>"
 \<s\>J'ai passé la soirée sur le Taïga Jet confiait Joffrey, 14 ans, c'est super.\</s\>
 \<s\>Il y a de l'ambiance, tous les ans j'y vais et à chaque fois c'est au top !\</s\> "
\</s\>
\<s\>Quant à Kévin, 8 ans, lui, il s'essayait près de sa maman à la machine à pinces.\</s\>

*Segmentation de la citation (4.2) obtenue en mars 2024.*

### 4.3 Améliorations partielles

Dans les 16 citations regroupées dans cette catégorie, la segmentation est de meilleure qualité qu'en 2019, mais des problèmes persistent. Dans six cas, le problème est lié à une étrangeté dans la ponctuation de la citation, telle que la présence de deux points successifs ou de guillemets englobants chevauchant deux phrases. Il semble difficile d'envisager de régler de tels problèmes sans risquer de dérégler ce qui fonctionne par ailleurs. Dans huit cas, le problème vient d'une absence d'indice. Dans les trois cas restants, des indices sont présents, mais ont mal été interprétés par CasFin. Dans le cas de la citation (4.3), la segmentation obtenue en 2024 règle en partie le problème de sursegmentation observée en 2019. *Chagny-Nevers* est bien intégré à la phrase principale, mais considéré comme une phrase incluse. En plus des versions 2019 et 2024, nous ajoutons ici la segmentation attendue.

(4.3) Jamais nous ne connûmes les splendeurs du wagon-restaurant : économies ! À Chagny, nous prenions une voie secondaire : Chagny-Nevers. Nous changions de train (attention aux bagages !) et montions dans des wagons autrement rustiques tirés par une lente machine poussive.

\<s\> Jamais nous ne connûmes les splendeurs du wagon-restaurant : économies ! \</s\>
\<s\> À Chagny, nous prenions une voie secondaire : \</s\>
\<s\> Chagny-Nevers. \</s\>
\<s\> Nous changions de train (attention aux bagages !) et montions dans des wagons autrement rustiques tirés par une lente machine poussive.\</s\>

*Segmentation de la citation (4.3) obtenue fin 2019.*

\<s\> Jamais nous ne connûmes les splendeurs du wagon-restaurant : économies ! \</s\>
\<s\> À Chagny, nous prenions une voie secondaire :
 \<s\> Chagny-Nevers. \</s\>
\</s\>
\<s\> Nous changions de train (attention aux bagages !) et montions dans des wagons autrement rustiques tirés par une lente machine poussive.\</s\>

*Segmentation de la citation (4.3) obtenue en mars 2024.*

\<s\> Jamais nous ne connûmes les splendeurs du wagon-restaurant : économies ! \</s\>
**\<s\>** À Chagny, nous prenions une voie secondaire : Chagny-Nevers. **\</s\>**
\<s\> Nous changions de train (attention aux bagages !) et montions dans des wagons autrement rustiques tirés par une lente machine poussive.\</s\>

*Segmentation attendue de la citation (4.3).*

## *4.4 Problèmes persistants*

Les cas de problèmes persistants sont au nombre de huit. Nous y avons observé trois cas de problèmes dans la source, deux cas d'absence d'indices et trois cas de mauvaise interprétation d'indices. Dans le cas de la citation (4.4) qui sert d'illustration ci-dessous, c'est la séquence *deux-point* suivit d'un chiffre qui fait échouer la segmentation et même si cet échec et sans conséquence sur les structures argumentales, il ne nous convient pas. Nous ne présentons pas ici la version 2019, mais uniquement la version 2024 et la segmentation attendue.

(4.4)   Dispositions statutaires communes aux corps des EPST (RMLR : 5112) Décret n° 2006-782 du 3 juillet 2006 modifiant le décret n° 48-1108 du 10 juillet 1948 portant classement hiérarchique des grades et emplois des personnels civils et militaires de l'État relevant du régime général des retraites Droit syndical (RMLR : 5233)

\<s\>Dispositions statutaires communes aux corps des EPST (RMLR :**\</s\>**
\<s\>5112) Décret n°2006-782 du 3 juillet 2006 modifiant le décret n°48-1108 du 10 juillet 1948 portant classement hiérarchique des grades et emplois des personnels civils et militaires de l'État relevant du régime général des retraites Droit syndical (RMLR :**\</s\>**
 **\<s\>**5233)\</s\>\</p\>

*Segmentation de la citation (4.4) obtenue en mars 2024.*

\<s\>Dispositions statutaires communes aux corps des EPST (RMLR : 5112) Décret n°2006-782 du 3 juillet 2006 modifiant le décret n°48-1108 du 10 juillet 1948 portant classement hiérarchique des grades et emplois des personnels civils et militaires de l'État relevant du régime général des retraites Droit syndical (RMLR : 5233)\</s\>

*Segmentation attendue de la citation (4.3).*

## *4.5 Nouveaux problèmes*

Les cas de nouveaux problèmes sont au nombre de onze. Nous y observons quatre cas de bizarreries dans la source et sept cas de mauvaises interprétations d'indices. Ces derniers sont en majorité liés à un problème d'interprétation du signe *point*. La segmentation attendue est ici identique à celle obtenue en 2019. Dans le cas de la citation (4.5) ci-dessous, c'est l'accumulation des parenthèses, de la majuscule en début du nom propre et du point d'initiale qui font échouer la segmentation, à deux reprises.

(4.5)   Dans « Après la vie », le cavaleur sauve et est sauvé par une camée en état de manque (Dominique B.), une accro à la morphine, femme d'un flic (Gilbert M.) qui lui fournit ses doses, la protège.

\<s\>Dans " Après la vie ", le cavaleur sauve et est sauvé par une camée en état de manque **(**
   **\<s\>**Dominique B**.\</s\>**
**)**, une accro à la morphine, femme d'un flic **(**
   **\<s\>**Gilbert M**.\</s\>**
**)** qui lui fournit ses doses, la protège.\</s\>

*Segmentation de la citation (4.5) obtenue en mars 2024.*

\<s\>Dans " Après la vie ", le cavaleur sauve et est sauvé par une camée en état de manque **(**Dominique B**.)**, une accro à la morphine, femme d'un flic **(**Gilbert M**.)** qui lui fournit ses doses, la protège.\</s\>

*Segmentation attendue de la citation (4.5) et obtenue en 2019.*

## 5. Évaluation chiffrée

Nous présentons ici une évaluation basée sur une sous-partie de notre corpus test, composée de 3 000 citations. Lors de la première phase de cette évaluation, nous avons regardé si la segmentation obtenue était correcte ou incorrecte et si les citations à segmenter contenaient des signes de ponctuation que notre modèle aurait pu interpréter comme des délimiteurs de segments : double-points, tirets, parenthèses, guillemets.

Comme vous pouvez le voir dans le tableau ci-dessous, le système retourne des résultats corrects dans 98,5% des cas et ne génère des phrases incluses que pour **3,5% des citations**[4].

| Total erreurs | Correct sans particularité | Correct avec inclusion | Correct sans inclusion mais délimiteurs | Total correct |
|---|---|---|---|---|
| 51 | 2291 | 109 | 551 | 2951 |
| 1,5% | 76,5% | 3,5% | 18,5% | 98,5% |

*Évaluation de la segmentation sur 3 000 citations.*

Dans un second temps, notre attention s'est portée sur la question des ruptures dans les relations syntaxiques. Nous avons alors repris les cas d'erreurs de segmentation et les cas de segmentation correcte impliquant des inclusions. Le tableau suivant montre que seuls 8% des erreurs persistantes provoquent des ruptures dans les relations syntaxiques entre éléments lexicaux. Il montre également que plus de la moitié des inclusions de phrases réalisées permettent de ne pas séparer des éléments lexicaux syntaxiquement liés.

---

[4] Le nombre total de citations du tableaau est égale à 3002 en raison de la présence de 2 citations contenant à la fois une erreur de segmentation et un inclusion de phrases pertinente.

| Erreurs avec ruptures | Erreurs sans rupture | Inclusions avec liens | Inclusions sans liens |
|---|---|---|---|
| 4 | 47 | 58 | 51 |

*Impact sur les relations syntaxiques dans les cas d'erreurs et d'inclusions*

## 6. Conclusion

Comme nous venons de le voir, la segmentation réalisée par la cascade de graphes **CasFin** améliore la qualité de la segmentation en phrases du corpus BEL-RL-fr, par rapport à celle obtenue à l'aide de **CasEN**. La possibilité d'inclure des phrases à l'intérieur de phrases englobantes répond de façon satisfaisante à des cas spécifiques d'arguments syntaxiques qui sont des phrases à part entière. Nous pouvons désormais envisager la distribution du corpus ainsi segmenté, auprès des lexicographes qui l'exploitent au quotidien, d'une part, et auprès de la communauté à travers la plateforme Ortolang, d'autre part.

La cascade CasFin est librement disponible sur le site TLN du laboratoire Lifat[5]. Elle va déjà servir dans la mise à disposition et à l'exploitation d'Orthocorpus, un corpus dédié à la langue de spécialité des orthophonistes.

## Bibliographie

---

[5] https://tln.lifat.univ-tours.fr/version-francaise/ressources/casfin